\documentclass[final]{anthology-ch} 

\usepackage{booktabs}
\usepackage{graphicx}
\usepackage{makecell}
\usepackage{multirow}
\usepackage{csquotes}
\usepackage{graphicx}
\usepackage{color, colortbl}
\usepackage[dvipsnames]{xcolor}
\usepackage{soul}

\definecolor{lightGray}{gray}{0.9}

\title{Says Who? Effective Zero-Shot Annotation of Focalization}

\author[1,2]{Rebecca M.\ M.\ Hicke}[
  orcid=0009-0006-2074-8376
]

\author[2]{Yuri Bizzoni}[
  orcid=0000-0002-6981-7903
]

\author[2]{Pascale Feldkamp}[
  orcid=0000-0002-2434-4268
]

\author[2,3,4]{Ross Deans Kristensen-McLachlan}[
  orcid=0000-0001-8714-1911
]

\affiliation{1}{Department of Computer Science, Cornell University, Ithaca, NY, USA}
\affiliation{2}{Center for Humanities Computing, Aarhus University, Aarhus, Denmark}
\affiliation{3}{Department of Linguistics, Cognitive Science, and Semiotics, Aarhus University, Aarhus, Denmark}
\affiliation{4}{TEXT - Center for the Contemporary Cultures of Text, Aarhus University, Aarhus, Denmark}


\pubyear{2025}
\pubvolume{1}
\pagestart{1}
\pageend{1}
\conferencename{Proceedings of Conference XXX}
\conferenceeditors{Editor1 Editor2}
\doi{00000/00000}  

\begin{document}

\maketitle

\begin{abstract}
Focalization describes the way in which access to narrative information is restricted or controlled based on the knowledge available to knowledge of the narrator. It is encoded via a wide range of lexico-grammatical features and is subject to reader interpretation. 
Even trained annotators frequently disagree on correct labels, suggesting this task is both qualitatively and computationally challenging. 
In this work, we test how well five contemporary large language model (LLM) families and two baselines perform when annotating short literary excerpts for focalization. 
Despite the challenging nature of the task, we find that LLMs show comparable performance to trained human annotators, with GPT-4o achieving an average F1 of 84.79\%.
Further, we demonstrate that the log probabilities output by GPT-family models frequently reflect the difficulty of annotating particular excerpts.
Finally, we provide a case study analyzing sixteen Stephen King novels, demonstrating the usefulness of this approach for computational literary studies and the insights gleaned from examining focalization at scale. 
\end{abstract}

\section{Introduction}

Narratology and narrative theory provide rich frameworks of complex textual phenomena which can be used to explain how narrative discourse is ordered and how readers process text \cite{emmott_narrative_1997, herman_storylogic_2002, sanford_emmott_2012}. Since its inception, narrative theory has been applied to a wide variety of domains, from news-writing \cite{ormen_news_2019} to political discourse \cite{schubert_forms_2010}. One particularly central notion is that of \textit{focalization}, or the way in which information is constrained by the narrator's knowledge or perspective \cite{genette_narrative_1990}. While focalization is often correlated with point-of-view, these phenomena are subtly different. Point of view (or perspective) describes the way a story is represented from a particular position or world view \cite{niederhoff_perspective-pov_2011}. Focalization, on the other hand, refers more narrowly to the way in which a reader's access to narrative information is selected or restricted based on the knowledge of a narrator or character in the text \cite{niederhoff_focalization_2011}. In other words, focalization takes into account not only who (or what) is narrating, but also how that narrative voice is situated in the context of the story \cite{scholes_narrative_2006} -- i.e., how information is disclosed. Analyses of focalization have shown promise in qualitative research for understanding both narrative structures and the impact of texts on readers.\footnote{While we focus on literature here, focalization is also used to describe film and other media \cite{deleyto_focalisation_1991}.} Existing research also demonstrates how narrative focalization affects readers' character identification and empathy, as well as their experience of immersivity in a narrative \cite{bruhns_internal_2024, andringa_effects_1996, jumpertz_empirical_2020}. As such, focalization both provides information on how narratives are constructed and serves as an intermediate step toward understanding more complex textual phenomena, such as affective reader response and identification. Moreover, the concept’s relevance extends beyond literary narratives: patterns of focalization may also inform how affective framing operates in news discourse or political communication, where perspective and information control shape readers’ emotional and ideological engagement \cite{sanders_linguistic_1993, otmakhova-frermann-2025-narrative}.

Recently, large language models (LLMs) have shown great promise for automating the annotation of many syntactic and semantic linguistic features \cite{thalken-etal-2023-modeling, hicke2024lions, soni2023grounding}. These automated annotations allow researchers to study linguistic and literary phenomena on previously infeasible scales. In addition, examining the ability of LLMs to perform these challenging tasks provides researchers with a fuller understanding of their capabilities, particularly their strengths and weaknesses when performing nuanced annotations of complex, real-world texts.

This paper therefore studies LLMs' ability to annotate literary texts for focalization. We consider scenarios in which there is little human-annotated data available in an effort to make this methodology accessible to those without the time, money, or expertise to produce such annotations. Specifically, we evaluate the ability of LLMs from five model families --- DistilBERT, RoBERTa, Flan-T5, Llama, and GPT --- and two baselines --- logistic regression and Naive Bayes --- to annotate excerpts from sixteen novels by Stephen King for focalization mode. We find that GPT-4o with a zero-shot prompt achieves high (F1 = $\sim$85\%) agreement with human consensus labels. Further, we determine that GPT-4o provides similar labelings when prompted multiple times on the same dataset and that it is resilient to prompt perturbations. Finally, we show that confidence values calculated from the log-probabilities output by GPT models frequently reflect the difficulty of annotating particular passages.

In addition, we demonstrate the usefulness of automated focalization annotations by analyzing the structure and flow of 16 novels by Stephen King, highlighting three outliers in this small corpus of his work. Considering the link between focalization and immersion in stories \cite{bruhns_internal_2024}, we compare these annotations with measures of sensory information in each novel, finding that King appeals primarily to different senses across different focalization modes.

\section{Related Works}

Considerable recent work in NLP and computational humanities has examined the ability of language models to perform a variety of text-annotation tasks. Much of this research has fine-tuned models for tasks like co-reference annotation \cite{hicke2024lions}, classifying legal reasoning \cite{thalken-etal-2023-modeling}, distinguishing between historical and contemporary novels \cite{bjerring2024literary}, recognizing spatial entities \cite{kababgi2024recognising}, and more \cite{bamman2024classification}.  With the advent of instruction-tuned LLMs like GPT and Llama, studies have further probed the ability of models to perform annotations in a zero or few-shot setting for tasks including identifying aspects of poetic form \cite{walsh2024sonnet}, story-telling \cite{antoniak-etal-2024-people}, character roles \cite{stammbach-etal-2022-heroes}, familial relationships \cite{pagel-etal-2024-evaluating}, sentiment \cite{rebora2023comparing}, genre \cite{kuzman2023chatgpt}, and more. While model performance varies across tasks, research has generally found that LLMs have a remarkable ability to perform nuanced literary and linguistic annotations even when given no or few examples. We build on this research by further exploring the capabilities of smaller, fine-tuned, and large, prompted LMs to annotate for focalization.

Despite the limitations of the term and its application \cite{nelles_getting_1990}, focalization has been useful in narrative analysis across both fiction and nonfiction texts. Generally, studies tend to adopt either the original taxonomy of \textcite{genette_narrative_1990}, or a revised approach such as that of \textcite{bal_narratology_2009} or \textcite{chatman_story_1980}. In the original taxonomy, \textcite{genette_narrative_1990} distinguishes between three main types of focalization: internal, external, and zero. The latter corresponds to what is also generally known as an omniscient narrator -- i.e., a type of narrator with knowledge of all events, as well as of the characters and their thoughts. In internal focalization, narration is restricted to what a single character knows, hears, or sees, while in external focalization, it is based on the outside observation of characters or events. 
Significantly, modes of focalization are not equivalent to narrative point-of-view; third-person narration in particular is frequently internally focalized. Indeed, this is a core feature of so-called \textit{free indirect style}, an important stylistic phenomenon in the history of English literature \cite{rundquist2017free}.

One of the major points of discussion since the term's introduction has been the distinction between external and zero focalization \cite{niederhoff_focalization_2011}. Further studies on focalization have demonstrated that it is a significant aspect of how readers process texts and how effects, such as suspense, affect, and identification, are generated in narrative text \cite{andringa_effects_1996, jumpertz_empirical_2020}.




Focalization was additionally mentioned as a task of interest in two recent surveys on computational narrative understanding \cite{piper-etal-2021-narrative, santana2023survey}, although there have been few attempts to automate its annotation. Researchers in narrative generation have proposed frameworks that incorporate focalization \cite{bae-etal-2011-towards, AkimotoOgata+2015+177+188} and some work has attempted to identify similar concepts such as stream-of-consciousness \cite{long-so-2016-turbulent} and types of speech (e.g.\ free indirect) \cite{Brunner2020ToBO}. However, none of these works use large, generative LMs or attempt to annotate directly for focalization. We are not aware of any existing research on automating focalization annotations at scale.

\section{Methods}

\subsection{Data}

\begin{table}
  \small
  \centering
  \begin{tabular}{lc}
     \toprule
     \textbf{Title} & \textbf{\# Excerpts} \\
     \midrule
     The Gunslinger & 430 \\
     \rowcolor{lightGray}
     The Girl Who Loved Tom Gordon & 516 \\
     Dolores Claiborne & 753 \\
     \rowcolor{lightGray}
     The Eyes Of The Dragon & 791 \\
     Misery & 821 \\
     \rowcolor{lightGray}
     Cujo & 907 \\
     The Green Mile & 967 \\
     \rowcolor{lightGray}
     The Dead Zone & 1,006 \\
     Firestarter & 1,103 \\
     \rowcolor{lightGray}
     Salem's Lot & 1,163 \\
     The Waste Lands & 1,298 \\
     \rowcolor{lightGray}
     Desperation & 1,436 \\
     Insomnia & 1,851 \\
     \rowcolor{lightGray}
     Wizard And Glass & 1,885 \\
     Needful Things & 1,898 \\
     \rowcolor{lightGray}
     The Stand & 3,295 \\
     \bottomrule
  \end{tabular}
  \caption{Total number of automatically annotated excerpts per book (cf. Section \ref{sec:5}).}
  \label{tab:bookLengths}
\end{table}

We study a corpus of 16 novels by Stephen King included in the Chicago Corpus\footnote{\url{https://textual-optics-lab.uchicago.edu/us_novel_corpus}} and published between 1975 and 1999 (listed in Appendix \ref{sec:corpusContents}). We focus on novels by King because he is known to use multiple modes of focalization within a novel, often to create suspense \cite{clasen_why_2020}. Moreover, King is known as an extraordinarily productive and versatile author, publishing over a long period and across genres connected to differing registers and narrative strategies \cite{van_cranenburgh_stylometric_2021, hye2023stephen, ketzan_anxiety_2024}. We may therefore expect King's use of focalization to vary across texts. In addition, studying the works of a single author allows us to determine whether focalization annotations can be used to identify structural outliers within a single author's oeuvre.

To prepare data for training and evaluating models, we split each novel into paragraphs and retain only excerpts of at least 50 words. We choose this method of segmenting the novels because paragraphs are salient textual units with breaks selected by the author and are thus narratively meaningful. In addition, studying multiparagraph textual units may lead to sections with shifting focalization being falsely labeled as zero focalization. For human annotation and validation, we draw sixteen excerpts from each novel, creating a dataset of 256 paragraphs in total. We then sample 50 further excerpts from six novels\footnote{\textit{Salem's Lot}, \textit{The Stand}, \textit{The Dead Zone}, \textit{Firestarter}, \textit{Cujo}, and \textit{The Gunslinger}} to create a minimal training dataset of 300 paragraphs. When annotating entire novels, we remove all excerpts containing front and back matter from the dataset. The remaining number of paragraphs annotated from each novel is displayed in Table \ref{tab:bookLengths}.

\begin{figure*}
    \centering
    \includegraphics[width=.65\linewidth]{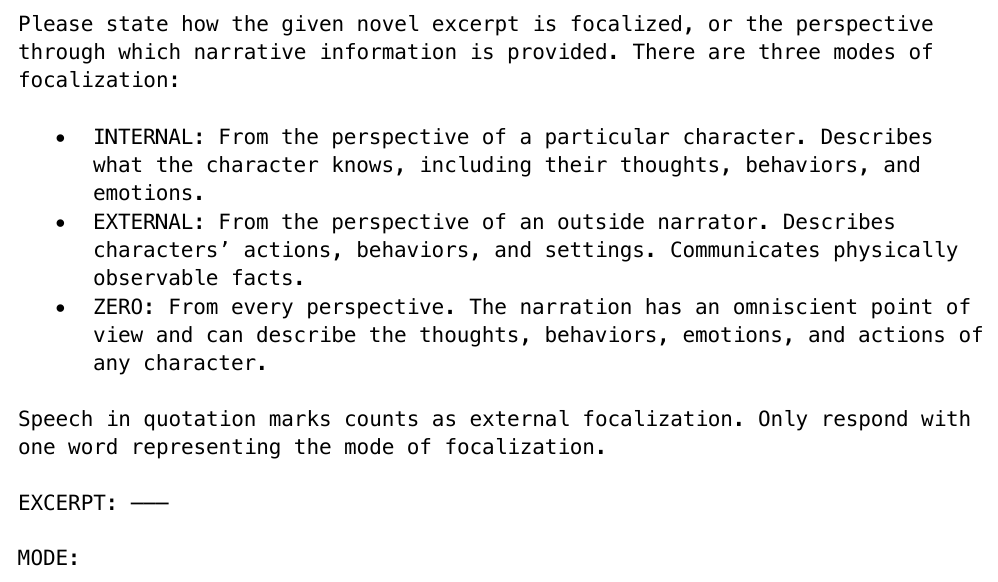}
    \caption{The prompt used for all zero-shot experiments. The definitions for each mode of focalization are adapted from those given in \textcite{chen2023}. The paragraph to be analyzed is inserted after the \texttt{EXCERPT:} heading.}
    \label{fig:modelPrompt}
\end{figure*}

\subsection{Human Annotation}

Each of the 256 excerpts in the evaluation dataset was annotated for focalization by three annotators. The first 96 samples were annotated by two native English speakers and one fluent speaker. Two of the annotators had post-secondary degrees in literature. The remaining 160 annotations were made by the two original annotators with post-secondary degrees in literature and by a third annotator, a fluent English speaker with a post-secondary degree in literature. In both cases, the annotators were instructed to work independently and did not have access to each other's annotations. Points of confusion were intentionally not discussed until after the initial annotation phase to produce a realistic measure of disagreement between informed human annotators. 

For guidance, the annotators were given the same prompt as the LLMs (Figure \ref{fig:modelPrompt}). The prompt includes a basic definition of focalization and short descriptions of each focalization mode, adapted from those in \textcite{chen2023}. One sentence in the prompt --- ``Speech in quotation marks counts as external focalization.'' --- was only added after the first 96 excerpts were annotated. Some disagreement between the human annotators thus arose from this point of confusion. Although the annotators had previous knowledge of some of the novels, all annotations were made only with the context provided by each excerpt.

After the initial annotation phase, two annotators discussed each disagreement to create a set of consensus labels for the initial 96 excerpts. After the second round of annotations, an external adjudicator (native English speaker, post-secondary degree in text linguistics) decided on final annotations for all disagreements. Krippendorf's alpha was used after each annotation batch to evaluate inter-annotator reliability.
Finally, the additional 300 training samples were labeled by two annotators from the second batch of evaluation annotation. Disagreements were discussed and settled by the annotators.

\subsection{Automated Annotation}

We then evaluated the ability of two baselines, three BERT-based models, five Flan-T5 models, four Llama 3 models, and three models from OpenAI's GPT family to annotate for focalization. These models were chosen to represent a range of computational power, accessibility, and architectures. A detailed model list is available in Table \ref{tab:modelF1s}.

The two baselines, a Naive Bayes classifier and logistic regression with word-count and TF-IDF features, were the least powerful but most accessible of the models we tested. These were included to determine whether focalization annotation is a trivial task that can be solved with only term frequencies.
Similarly, DistilBERT, the two RoBERTA models, and the three smallest Flan-T5 models were evaluated to determine whether relatively lightweight models fine-tuned on a minimal dataset were sufficiently powerful to generate the annotations of interest. The two largest Flan-T5 models and the four Llama 3 models are larger and more powerful than the baselines, but are open source and accessible at no cost. Finally, the GPT models are the largest and the least accessible of the tested models as they require a per-token cost to query. These models assessed whether large, powerful models could identify focalization in a zero-shot setting.

We implemented the Naive Bayes and logistic regression classifiers and vectorizers using the \texttt{sklearn} package and trained them on an additional 300-sample training dataset. The DistilBERT model, RoBERTa models, and the three smallest Flan-T5 models were accessed and fine-tuned using HuggingFace. From the 300-sample training dataset, two excerpts labeled as zero focalization, eight labeled as external, and ten labeled as internal were included in the validation dataset, and the remaining 280 samples were used for training. Further fine-tuning parameters can be found in Appendix \ref{sec:finetuningParams}. 
The two largest Flan-T5 models and the Llama 3 models were also accessed via HuggingFace. For the Llama models, ``You are a helpful assistant.'' was passed as a generic system message. For each excerpt, the zero-shot prompt in Figure \ref{fig:modelPrompt} was passed to the model with the excerpt of interest appended.

Finally, OpenAI's API was used to access the GPT models. Each time a GPT model was queried, we again passed ``You are a helpful assistant.'' as a system message and the zero-shot prompt with the excerpt appended as a user message. The default parameter values were used for all but two parameters: nuclear sampling (\texttt{top\_p}) and log-probabilities (\texttt{logprob}). \texttt{top\_p} was set to 0.1, which forced the model to select tokens from only the top 10\% of the probability mass. This was intended to optimize the model's performance for accuracy. \texttt{logprob} was set to \texttt{True} so the model output the log-probabilities for each token. 

The log-probability values were then used to create a proxy `confidence' metric for each annotation. Specifically, the log-probability for the first token of each annotation was converted into a probability $p$ ($p=e^{l}$, where $l$ is the log probability). Since no GPT model provided an annotation outside of the expected answer set (internal, external, or zero) and the first token of each of these modes is unique, the reported probability of the first token was interpreted as model confidence.

All baselines and models were used to annotate the evaluation dataset of 256 excerpts. The non-baseline models were asked to annotate the same texts three times, and the performance was averaged over all runs. Further, to test the highest-performing model's resistance to prompt variations, we created five alternate versions of the original prompt (Appendix \ref{sec:promptVariants}) and evaluated the model using each of these prompts. Finally, we used this same model to annotate all $\geq$ 50 word excerpts from each of the 16 novels in our subcorpus.

\section{Annotation Accuracy and Agreement}

\subsection{Human Performance}
Human annotators achieved mid-to-low inter-annotator reliability on this task (first round: $\alpha$ = 0.55, second round: $\alpha$ = 0.65), demonstrating its difficulty.\footnote{Note that this is not uncommon for annotations of literary texts, which often exhibit comparable inter-annotator reliability scores, for example, in event type annotation \cite{vauth_event_2022}.} The annotators reported several primary challenges. First, they found it difficult to establish whether environmental information was communicated through a given character's perspective or through an external observer. For example, in the following quote, it is challenging to determine whether the use of ``could see'' is internally focalized through Trisha or externally through an observer: 

\begin{displayquote}
    ``Even in this channel, she was forced to clamber over one downed tree. It had fallen just recently, and `fallen' was really the wrong word. Trisha could see more slash-marks in its bark, and (...) she could see how fresh and white the wood of the stump was.''\footnote{\textit{The Girl Who Loved Tom Gordon} by Stephen King}
\end{displayquote}

After discussion, the annotators agreed to label such instances as internally focalized, even if the information was externally observable, as long as there was some keyword indicating that it was perceived by a particular character (e.g.\ saw, heard). The annotators were also challenged by paragraphs where thoughts clearly belonging to a particular character were given as narration, as in the following quote:

\begin{displayquote}
    ``Aye, as they said in these parts. If the boy had had the impertinence to begin an affair with the Mayor’s gilly-in-waiting, and the incredible slyness to get away with it, what did that do to Jonas’s picture of three In-World brats who could barely find their own behinds with both hands and a candle?''\footnote{\textit{Wizard and Glass} by Stephen King}
\end{displayquote}

The annotators also agreed to label such examples as internally focalized. Overall, despite initial disagreements, the annotators reached consensus on labels for each excerpt in the evaluation dataset.

\begin{table*}[t]
    \small
  \centering
  \begin{tabular}{lcccccc}
    \toprule
        & \multicolumn{1}{c}{\textbf{Internal}} & \multicolumn{1}{c}{\textbf{External}} & \multicolumn{1}{c}{\textbf{Zero}} & \multicolumn{3}{c}{\textbf{Overall}} \\
        \cmidrule(lr){2-2} \cmidrule(lr){3-3} \cmidrule(lr){4-4} \cmidrule(lr){5-7}
        \textbf{Model} & F1 & F1 & F1 & Precision & Recall & F1 \\
        \midrule
        Logistic Regression* & 81.23 & 62.86 & 0.0 & 69.96 & 73.83 & 71.84 \\
        Naive Bayes* & 83.05 & 39.56 & 0.0 & 70.86 & 73.05 & 66.78 \\
        \midrule
        DistilBERT$^{\dag}$* & 85.16 & 61.07 & 0.0 & 73.94 & 77.60 & 74.01 \\
        RoBERTa Base$^{\dag}$* & 85.07 & 64.53 & 0.0 & 73.60 & 77.86 & 74.88 \\
        RoBERTa Large$^{\dag}$* & 86.06 & 71.65 & 0.0 & 75.54 & 79.82 & 77.47 \\
        \midrule
        Flan-T5 Small$^{\dag}$* & 69.56 & 30.59 & 0.0 & 55.61 & 59.64 & 55.25 \\
        Flan-T5 Base$^{\dag}$* & 82.34 & 23.28 & 0.0 & 73.49 & 71.09 & 61.92 \\
        Flan-T5 Large$^{\dag}$* & 86.47 & 67.31 & 0.0 & 75.97 & 79.82 & 76.58 \\
        Flan-T5 XL$^{\dag}$ & 77.92 & 30.19 & 19.05 & 61.03 & 65.62 & 61.84 \\
        Flan-T5 XXL$^{\dag}$ & 81.87 & 33.33 & 22.22 & 70.37 & 69.14 & 65.52 \\
        \midrule
        Llama 3.2 1b$^{\dag}$ & 0.0 & 3.28 & 4.55 & 3.12 & 0.91 & 1.13 \\
        Llama 3.2 3b$^{\dag}$ & 78.49 & 31.19 & 22.38 & 62.51 & 66.80 & 62.67 \\
        Llama 3.1 8b$^{\dag}$ & 63.39 & 55.39 & 0.0 & 75.69 & 57.68 & 57.76 \\
        Llama 3.3 70b$^{\dag}$ & 86.34 & 73.88 & 13.02 & 80.95 & 79.30 & 78.97 \\
        \midrule
        GPT-3.5-turbo$^{\dag}$ & 75.02 & 64.19 & 29.82 & 76.56 & 68.49 & 69.63 \\
        GPT-4-turbo$^{\dag}$ & 88.33 & 71.79 & 32.55 & 81.21 & 82.03 & 80.82 \\
        GPT-4o$^{\dag}$ & \textbf{88.73} & \textbf{84.78} & \textbf{36.16} & \textbf{86.63} & \textbf{84.64} & \textbf{84.79} \\
        \bottomrule
    \end{tabular}
  \caption{Evaluations of model performance on focalization annotation. Overall scores are weighted by class size. Values reported for models marked with a dagger ($^{\dag}$) are the averages over three runs, and those marked with an asterisk (*) were fine-tuned on 300 samples. All other models were prompted using the prompt in Figure \ref{fig:modelPrompt}. The values reported for Naive Bayes and logistic regression are the highest performing over several input feature variants; further results for these models are reported in Appendix \ref{sec:baselineResults}.}
  \label{tab:modelF1s}
\end{table*}

\subsection{Model Performance}
We evaluated the annotations produced by all computational baselines and LLMs against the consensus labels (Table \ref{tab:modelF1s}). GPT-4o was the highest performing model, with an overall F1 score of 84.79\%. All other models and baselines except for Llama 3.2 1b achieved considerably above-random performance, with F1 scores ranging from 55.25\% (Flan-T5 small) to 80.82\% (GPT-4-turbo). However, none performed on par with GPT-4o.

All models except Llama 3.2 1b achieved the highest F1 score for internal focalization, suggesting that it was the easiest mode to identify. In contrast, we find that the models nearly all performed the worst when annotating for zero focalization; in fact, of the seventeen tested models, nine failed to label any excerpts as zero-focalized. This may indicate the difficulty of identifying zero focalization, or may be a side-effect of class imbalance in the evaluation dataset.


\begin{table}
  \small
  \centering
  \begin{tabular}{lcccccc}
     \toprule
     & \multicolumn{2}{c}{\textbf{GPT 3.5-turbo}} & \multicolumn{2}{c}{\textbf{GPT 4-turbo}} & \multicolumn{2}{c}{\textbf{GPT 4o}} \\
     \cmidrule(lr){2-3} \cmidrule(lr){4-5} \cmidrule(lr){6-7}
     & Agree & Disagree & Agree & Disagree & Agree & Disagree \\
     \midrule
     \textbf{Humans} & 0.59 ($\pm$0.1) & 0.54 ($\pm$0.2) & 0.98 ($\pm$0.1) & 0.96 ($\pm$0.1) & 0.96 ($\pm$0.1) & 0.91 ($\pm$0.1) \\
     \textbf{GPT Models} & 0.61 ($\pm$0.1) & 0.52 ($\pm$0.1) & 0.98 ($\pm$0.1) & 0.96 ($\pm$0.1) & 0.98 ($\pm$0.1) & 0.91 ($\pm$0.1) \\
     \textbf{4o} & 0.58 ($\pm$0.1) & 0.56 ($\pm$0.1) & 0.98 ($\pm$0.1) & 0.89 ($\pm$0.2) & 0.96 ($\pm$0.1) & 0.72 ($\pm$0.2) \\
     \textbf{4o (Prompts)} & 0.60 ($\pm$0.1) & 0.53 ($\pm$0.1) & 0.98 ($\pm$0.1) & 0.95 ($\pm$0.1) & 0.99 ($\pm$0.0) & 0.85 ($\pm$0.2) \\
     \bottomrule
  \end{tabular}
  \caption{The average of each model's confidence values when a subset of annotators (left column) agree or disagree. Standard deviations are given in parentheses. The differences in means are significant at $\alpha = 10^{-2}$ except for the confidence values of GPT 3.5-turbo when GPT 4o agrees and disagrees.}
  \label{tab:avgConfs}
\end{table}

Interestingly, the difference in F1 scores for the GPT models suggests that changes made between GPT-3.5 (first released March 15, 2022) and the GPT-4 models (first released March 14, 2023) significantly improved their ability to perform focalization annotation, and perhaps literary annotation more broadly. This difference in performance was reflected in the models' reported confidence values; GPT-3.5-turbo was, on average, less confident in its predictions than other models (57.6\% as compared to 97.2\% (GPT-4-turbo) and 95.0\% (GPT-4o)). This provides some evidence that the confidence values reflect model accuracy. An ANOVA test suggests we can reject the null-hypothesis, namely that the confidence values from all three models are drawn from the same distribution ($p < 0.001$\footnote{$p$-value reported as $5.7\text{x}10^{-208}$}). The three GPT models also have low inter-annotator reliability ($\alpha$ = 0.47).

However, GPT-4o demonstrated very high consistency across three model runs ($\alpha$ = 0.94), with F1 scores only ranging from 84.1\% to 84.7\%. In this case, an ANOVA test does not refute the null hypothesis ($p \approx 0.90$), indicating no statistically significant variance between the confidence values produced by each model. This further suggests that GPT-4o can apply a consistent paradigm for annotating texts for focalization, achieving relatively high accuracy in alignment with human consensus annotations.

Furthermore, we found that GPT-4o was not particularly sensitive to variations in prompts. The model achieved a Krippendorf's alpha of 0.74 across all six prompt variants. The F1 scores for the variant annotations ranged from 79.8\% (variant \#3) to 84.5\% (variant \#5). Notably, Variant \#5 provides the least explanation of all the tested prompts. Yet, it achieves nearly as high performance as the base prompt, which may indicate that GPT-4o had an understanding of focalization from pre-training. This resilience to prompt perturbations is notable given past research \cite{abraham2024promptselectionmattersenhancing, coyne2023analyzingperformancegpt35gpt4, lu-etal-2022-fantastically, zhao2021calibrate, gan-mori-2023-sensitivity}, and again suggests that GPT-4o is a reliable annotator of focalization.

Finally, we found that the confidence values produced by the GPT models corresponded to outside signals of the difficulty of annotating an excerpt. On average, texts that appeared to be more difficult to annotate received lower average confidence values (Table \ref{tab:avgConfs}). This is true for texts where humans disagreed, the three GPT models disagreed, the GPT-4o runs disagreed, and GPT-4o with prompt perturbations disagreed. These differences are relatively small but are largely significant, suggesting that the confidence values correlate to some degree with the ambiguity of an annotation.

\section{Focalization at Scale}
\label{sec:5}

To validate the usefulness of LLM annotations for focalization, we examine the distribution of focalization modes across the subcorpus of 16 Stephen King novels. Specifically, we examine whether the focalization annotations enable us to identify structural outliers in King's works and to show the association between focalization mode and the prominence of sensory descriptors.

\subsection{Comparing Novel Structures}

\begin{table*}
  \small
  \centering
  \begin{tabular}{lcccc}
    \hline
    \textbf{Novel} & \textbf{\% Internal} & \textbf{\% External} & \textbf{\% Zero} \\
    \hline
    The Girl Who Loved Tom Gordon & 84.3 & 13.0 & 2.7 \\
    \rowcolor{lightGray}
    Dolores Claiborne & 81.4 & 18.2 & 0.4 \\
    Cujo & 68.5 & 25.3 & 6.3 \\
    \rowcolor{lightGray}
    Misery & 66.5 & 30.5 & 3.1 \\
    The Green Mile & 62.7 & 32.7 & 4.7 \\
    \rowcolor{lightGray}
    Insomnia & 60.4 & 35.6 & 4.1 \\
    Firestarter & 59.6 & 34.3 & 6.2 \\
    \rowcolor{lightGray}
    Desperation & 59.5 & 37.6 & 2.9 \\
    Needful Things & 58.9 & 36.2 & 4.9 \\
    \rowcolor{lightGray}
    The Stand & 56.9 & 38.7 & 4.4 \\
    Wizard and Glass & 55.5 & 38.5 & 6.1 \\
    \rowcolor{lightGray}
    The Waste Lands & 55.2 & 38.4 & 6.3 \\
    The Gunslinger & 52.4 & 38.4 & 9.3 \\
    \rowcolor{lightGray}
    The Eyes of the Dragon & 48.7 & 23.6 & 27.7 \\
    The Dead Zone & 47.9 & 46.4 & 5.7 \\
    \rowcolor{lightGray}
    Salem's Lot & 45.7 & 49.6 & 4.7 \\
    \bottomrule
  \end{tabular}
  \caption{The percentage of paragraphs with $\geq 50$ words from each novel that were annotated as internally, externally, or zero focalized.}
  \label{tab:percentModes}
\end{table*}


From the annotations created by GPT-4o, we find that the typical King novel is primarily internally focalized, with some externally and few zero focalized excerpts distributed regularly throughout its course.
However, some novels deviate from this basic structure. In particular, two novels --- \textit{The Girl Who Loved Tom Gordon} and \textit{Dolores Claiborne} --- have a much higher percentage of internally focalized paragraphs (Table \ref{tab:percentModes}). There are clear reasons why this is likely to be true for both texts. \textit{The Girl Who Loved Tom Gordon} follows a young girl lost alone in the woods, and thus focalizes much of the narrative through her eyes. Additionally, \textit{Dolores Claiborne} is written as an almost stream-of-consciousness narrative from the perspective of the titular character, and thus is again primarily internally focalized through that character.

Another outlier novel is \textit{The Eyes of the Dragon}, which has a much higher percentage of zero focalized excerpts than any other novel (Table \ref{tab:percentModes}). This is likely because the novel has a fairy-tale-esque omniscient narrator, who often describes the perspectives of multiple or groups of characters at once, as in the following quote:

\begin{displayquote}
    ``He had been allowed to marry late because he had met no woman who pleased his fancy, and because his mother, the great Dowager Queen of Delain, \textbf{\textit{had seemed immortal to Roland and to everyone else —-- and that included her}}.''\footnote{\textit{The Eyes of the Dragon} by Stephen King}
\end{displayquote}

Overall, we find that focalization annotations enable large-scale comparisons of novels, highlighting structural differences even within the works of a single author.

\subsection{Links to Linguistic Features}

\begin{table}[ht]
\small
\centering
\begin{tabular}{l r r r}
\toprule
\textbf{Sense} & \textbf{Internal} & \textbf{External} & \textbf{Zero} \\
\midrule
Taste      & $^{*}$0.70 & $^{*}$-0.75 & -0.06 \\
Interoception  & $^{*}$0.54 & $^{*}$-0.65 & 0.06 \\
Touch         & $^{*}$0.56 & -0.34 & -0.46 \\
Smell      & $^{*}$0.51 & -0.44 & -0.21 \\
Sound       & -0.25 &  0.20 &  0.14 \\
Sight         & -0.21 & 0.34 & -0.17 \\
\bottomrule
\end{tabular}
\caption{Pearson's R correlations between sensorial information and focalization modes for each novel by King. All starred correlations have significance values $p<0.05$.}
\label{tab:sensoryCorrs}
\end{table}

Finally, we compared the annotations produced by GPT-4o to measures of other linguistic features. Scholars have previously hypothesized that internal focalization may be linked to a more \textit{immersive} style, or one that attempts to pull readers into the reality of the text and make it ``feel real'' \cite{allan2017enargeia, jacobs2017immersion, jumpertz_empirical_2020}. To explore the validity of this hypothesis, we examine a connection between focalization and sensory information, which we take as a proxy for immersion. While texts that are externally or zero focalized can contain sensory information, we seek to determine whether internally focalized texts are more likely to contain such sensory descriptors.

To test this hypothesis, we use the Lancaster Sensorimotor Lexicon \cite{lynott2020lancaster}, which contains sensorimotor strength values for $\sim$40,000 words across six perceptual axes: touch, hearing, smell, taste, vision, and interoception. For each of the 16 novels by King, we calculate the summed sensorimotor strength along each axis and standardize the value by the total number of words from the lexicon 
in the novel.\footnote{This encompasses nearly every word in the novel except for some function words like `of' or `is.'} Finally, we examine the relationship between the mean sensorimotor strength along each axis for each novel and the percentage of excerpts from the novel that were labeled as internally or externally focalized (Table \ref{tab:sensoryCorrs}).

We find that there are relatively strong positive correlations between the percentage of internal focalized paragraphs in a novel and all perceptual axes except for sound and sight. There are particularly strong positive correlations between internal focalization and interoception (0.54) and taste (0.70), which may be less perceptible to an `external' observer. Crucially, the opposite appears to be true for external focalization, which is negatively correlated with all senses except sound and sight. 
The correlations between the sensory values and the percentage of zero-focalized excerpts were generally weaker and not statistically significant. However, we found a notable negative correlation between zero focalization and touch. This may suggest that omniscient narrators focus less on characters' haptic sensory experiences, employ a language that is more reflective than descriptive \cite{gittel_neither_2024}, or may be an artefact of the skewed distribution of zero focalization values.

Overall, it appears that King is more likely to focus on the senses of smell, taste, touch, and interoception in internally focalized text and sound and sight in externally focalized text. This suggests that internally focalized texts may be more immersive, as they contain a greater variety of sensory descriptors, or at least that different forms of immersivity characterize King's uses of internal and external focalization. Whether this is true of focalization generally, rather than specific to King's writing, remains an open question.

\section{Conclusion}

In this paper, we demonstrated the usability of LLMs for a well-defined literary annotation task --- namely, the presence of focalization in narrative discourse. We find that performance on this task varies across different model architectures and sizes, but that, broadly speaking, larger, prompted LLMs outperform smaller models fine-tuned on a minimal dataset. Perhaps unexpectedly, we find that, while the GPT family models are not terribly consistent with each other, the GPT-4o model is reliable across multiple runs with the same prompt and within prompt variants.

Overall, we conclude that zero-shot GPT-4o is an effective and reliable annotator of focalization and that other models, including Llama 3.3 70b, are promising, if not as reliable.  In particular, the resilience of GPT-4o across prompt variations suggests that it consistently applies elements of literary theory and narratology. This indicates that LLMs hold great promise as annotators for a variety of literary tasks that require a subtle understanding of semantic and syntactic content, and that they will likely prove useful for even more abstract literary annotations.

\section*{Acknowledgements}
We would like to thank Mathias Clasen for his help brainstorming and formulating the research questions addressed in this paper. Additionally, part of the computation for this project was performed on the UCloud interactive HPC system, managed by the eScience Center at the University of Southern Denmark.

\printbibliography

\appendix
\clearpage

\section{Corpus Contents}
\label{sec:corpusContents}

{\centering
\begin{tabular}{lc}
\toprule
\textbf{Title} & \textbf{Publication Date} \\
\midrule
Salem’s Lot & 1975 \\
\rowcolor{lightGray}
The Stand & 1978 \\
The Dead Zone & 1979 \\
\rowcolor{lightGray}
Firestarter & 1980 \\
The Waste Lands & 1981 \\
\rowcolor{lightGray}
Cujo & 1982 \\
The Gunslinger & 1982 \\
\rowcolor{lightGray}
The Eyes of the Dragon & 1987 \\
Misery & 1987 \\
\rowcolor{lightGray}
Needful Things & 1991 \\
Dolores Claiborne & 1992 \\
\rowcolor{lightGray}
Insomnia & 1994 \\
Desperation & 1996 \\
\rowcolor{lightGray}
The Green Mile & 1997 \\
Wizard and Glass & 1997 \\
\rowcolor{lightGray}
The Girl Who Loved Tom Gordon & 1999 \\
\bottomrule
\end{tabular}
}

\section{Fine-Tuning Parameters}
\label{sec:finetuningParams}

{\centering
\begin{tabular}{lc}
\toprule
\textbf{Parameter} & \textbf{Value} \\
\midrule
Evaluation Strategy & \texttt{epoch} \\
\rowcolor{lightGray}
Save Strategy & \texttt{epoch} \\
Learning Rate & $2\times10^{-5}$ \\
\rowcolor{lightGray}
Weight Decay & 0.01 \\
\# Train Epochs & 5 \\
\rowcolor{lightGray}
Load Best Model at End & \texttt{True} \\
\bottomrule
\end{tabular}}

\section{Prompt Variants}
\label{sec:promptVariants}

Text in red indicates where changes from the original prompt (Figure \ref{fig:modelPrompt}) occurred. \\
\\
\textbf{Variant \#1} \\
Please state how the given novel excerpt is focalized\textcolor{BrickRed}{\st{, or the perspective through which narrative information is provided}}. There are three modes of focalization:

\begin{itemize}
    \item INTERNAL: From the perspective of a particular character. Describes what the character knows, including their thoughts, behaviors, and emotions.
    \item EXTERNAL: From the perspective of an outside narrator. Describes characters’ actions, behaviors, and settings. Communicates physically observable facts.
    \item ZERO: From every perspective. The narration has an omniscient point of view and can describe the thoughts, behaviors, emotions, and actions of any character.
\end{itemize}
Speech in quotation marks counts as external focalization. Only respond with one word representing the mode of focalization. \\
\\
\textbf{Variant \#2}\\
\textcolor{BrickRed}{Please state the perspective through which narrative information is provided in the given novel excerpt.} There are three modes\textcolor{BrickRed}{\st{ of focalization}}:

\begin{itemize}
    \item INTERNAL: From the perspective of a particular character. Describes what the character knows, including their thoughts, behaviors, and emotions.
    \item EXTERNAL: From the perspective of an outside narrator. Describes characters’ actions, behaviors, and settings. Communicates physically observable facts.
    \item ZERO: From every perspective. The narration has an omniscient point of view and can describe the thoughts, behaviors, emotions, and actions of any character.
\end{itemize}
Speech in quotation marks counts as external\textcolor{BrickRed}{\st{ focalization}}. Only respond with one word representing the mode\textcolor{BrickRed}{\st{ of focalization}}.\\
\\
\textbf{Variant \#3} \\
Please state how the given novel excerpt is focalized, or the perspective through which narrative information is provided. There are three modes of focalization:

\begin{itemize}
    \item INTERNAL: \textcolor{BrickRed}{\st{From the perspective of a particular character.}} Describes what the character knows, including their thoughts, behaviors, and emotions.
    \item EXTERNAL: \textcolor{BrickRed}{\st{From the perspective of an outside narrator.}} Describes characters’ actions, behaviors, and settings. Communicates physically observable facts.
    \item ZERO: \textcolor{BrickRed}{\st{From every perspective.}} The narration has an omniscient point of view and can describe the thoughts, behaviors, emotions, and actions of any character.
\end{itemize}
Speech in quotation marks counts as external focalization. Only respond with one word representing the mode of focalization. \\
\\
\textbf{Variant \#4} \\
Please state how the given novel excerpt is focalized, or the perspective through which narrative information is provided. There are three modes of focalization:

\begin{itemize}
    \item INTERNAL: From the perspective of a particular character. \textcolor{BrickRed}{\st{Describes what the character knows, including their thoughts, behaviors, and emotions.}}
    \item EXTERNAL: From the perspective of an outside narrator. \textcolor{BrickRed}{\st{Describes characters’ actions, behaviors, and settings. Communicates physically observable facts.}}
    \item ZERO: From every perspective. \textcolor{BrickRed}{\st{The narration has an omniscient point of view and can describe the thoughts, behaviors, emotions, and actions of any character.}}
\end{itemize}
Speech in quotation marks counts as external focalization. Only respond with one word representing the mode of focalization. \\
\\
\textbf{Variant \#5} \\
Please state how the given novel excerpt is focalized, or the perspective through which narrative information is provided. There are three modes of focalization: \textcolor{BrickRed}{INTERNAL, EXTERNAL, ZERO}

Speech in quotation marks counts as external focalization. Only respond with one word representing the mode of focalization. \\
\\

\section{Baseline Results}
\label{sec:baselineResults}

\begin{table*}[h]
    \small
  \centering
  \begin{tabular}{cccccccc}
    \toprule
        & & \multicolumn{1}{c}{\textbf{Internal}} & \multicolumn{1}{c}{\textbf{External}} & \multicolumn{1}{c}{\textbf{Zero}} & \multicolumn{3}{c}{\textbf{Overall}} \\
        \cmidrule(lr){3-3} \cmidrule(lr){4-4} \cmidrule(lr){5-5} \cmidrule(lr){6-8}
        \textbf{Features} & \textbf{Ngrams} & F1 & F1 & F1 & Precision & Recall & F1 \\
        \midrule
        \multicolumn{8}{c}{Logistic Regression} \\
        \midrule
        Count & 1 & 78.86 & 56.94 & 0.0 & 67.42 & 69.92 & 68.84 \\
        Count & 1--2 & 81.23 & \textbf{62.86} & 0.0 & 69.96 & 73.83 & \textbf{71.84} \\
        Count & 1--3 & 82.19 & 60.15 & 0.0 & 69.64 & \textbf{74.22} & 71.76 \\
        TF-IDF & 1 & \textbf{83.21} & 46.46 & 0.0 & \textbf{70.30} & 73.83 & 68.75 \\
        TF-IDF & 1--2 & 81.93 & 26.51 & 0.0 & 68.65 & 70.70 & 62.51 \\
        TF-IDF & 1--3 & 81.15 & 17.72 & 0.0 & 65.57 & 69.14 & 59.61 \\
        \midrule
        \multicolumn{8}{c}{Naive Bayes} \\
        \midrule
        Count & 1 & \textbf{83.05} & \textbf{39.56} & 0.0 & 70.86 & \textbf{73.05} & \textbf{66.78} \\
        Count & 1--2 & 82.58 & 25.32 & 0.0 & \textbf{74.48} & 71.48 & 62.63 \\
        Count & 1--3 & 81.99 & 18.42 & 0.0 & 73.90 & 70.31 & 60.37 \\
        TF-IDF & 1 & 80.65 & 0.0 & 0.0 & 45.67 & 67.58 & 54.50 \\
        TF-IDF & 1--2 & 80.65 & 0.0 & 0.0 & 45.67 & 67.58 & 54.50 \\
        TF-IDF & 1--3 & 80.65 & 0.0 & 0.0 & 45.67 & 67.58 & 54.50 \\
        \bottomrule
    \end{tabular}
  \caption{Evaluations of model performance on focalization annotation for all Naive Bayes and Logistic Regression models. The highest value in each column is bolded.}
  \label{tab:baselineF1s}
\end{table*}

\end{document}